\documentclass[letterpaper, 10pt, conference]{ieeeconf}      

\IEEEoverridecommandlockouts                              

\usepackage{cite}
\usepackage{amsmath,amssymb,amsfonts}
\usepackage{algorithmic}
\usepackage{graphicx}
\usepackage{comment}
\usepackage[dvipsnames]{xcolor}
\usepackage{fancyhdr}

\title{\LARGE \bf Fairness-Sensitive Policy-Gradient Reinforcement Learning for Reducing Bias in Robotic Assistance
}

\author{Jie Zhu$^{1,2}$, Mengsha Hu$^{1,3}$, Xueyao Liang$^{1}$, Amy Zhang$^{1}$, Ruoming Jin$^{3}$, Rui Liu$^{1*}$
\thanks{$^{1}$ is with the Cognitive Robotics and AI Lab (CRAI), College of Aeronautics and Engineering, Kent State University, Kent, OH 44240, USA. $^{2}$ is with Department of Computer Science George Washington University. $^{3}$ is with Department of Computer Science Kent State University. $^{*}$ Rui Liu is the corresponding author, email: ruiliu.robotics@gmail.com.}%
}

\graphicspath{{img/}}
\begin{document}

\maketitle
\thispagestyle{fancy} 
\renewcommand{\headrulewidth}{0pt} 
\renewcommand{\footrulewidth}{0pt}
\pagestyle{fancy}
\cfoot{\thepage}


\begin{abstract}
Robots assist humans in various activities, from daily living public service (e.g., airports and restaurants), and to collaborative manufacturing. However, it is risky to assume that the knowledge and strategies robots learned from one group of people can apply to other groups. 
The discriminatory performance of robots will undermine their service quality for some people, ignore their service requests, and even offend them. Therefore, it is critically important to mitigate bias in robot decision-making for more fair services. In this paper, we designed a self-reflective mechanism -- Fairness-Sensitive Policy Gradient Reinforcement Learning (FSPGRL), to help robots to self-identify biased behaviors during interactions with humans. FSPGRL identifies bias by examining the abnormal update along particular gradients and updates the policy network to support fair decision-making for robots. To validate FSPGRL's effectiveness, a human-centered service scenario, "A robot is serving people in a restaurant," was designed. A user study was conducted; 24 human subjects participated in generating 1,000 service demonstrations. Four commonly-seen issues \textit{"Willingness Issue," "Priority Issue," "Quality Issue," "Risk Issue"} were observed from robot behaviors. By using FSPGRL to improve robot decisions, robots were proven to have a self-bias detection capability for a more fair service. We have achieved the suppression of bias and improved the quality during the process of robot learning to realize a relatively fair model.

\end{abstract}


\section{INTRODUCTION}
As sensor and artificial intelligence technologies advance, robots now are able to assist humans in various activities from daily living, to public service, and manufacturing. For example, humanoid robots in hotels help guests to greet people, carry luggage, deliver packages to guest rooms, and clean rooms \cite{su13084431}; armed robots in collaborative manufacturing handover parts to workers and assist workers to control processing safety \cite{robotics8040100}. 

AI is developing fast and serving wide application areas, while fairness in AI attracts social attention as people are getting more concerned about safety, ethics, and accessibility issues during AI tool implementations. In 2016, the COMPAS software revealed algorithmic bias in deciding if set free an offender or not \cite{Propublica.2016}. Blacks are nearly twice as likely as whites to be labeled as higher risk, even though they do not actually re-offend. Subsequent research pointed out how unfair algorithms can discriminate against certain groups in hiring and lending decisions \cite{10.1145/3457607,doi10.48550, https://doi.org/10.1111/1475-3995.00375}. For example, One resume screening company discovered that being named "Jared" and participating in high school lacrosse were strong indicators of success in their model. To improve the safety and fairness of AI models, bias mitigation techniques for AI have been developed. Work \cite{DBLP:journals/corr/abs-2107-06720,pmlr-v139-correa21a} investigated optimal fair raking algorithm to solve online selection problems where decisions are often biased like hiring and credit risk estimating.

Inspired by the AI breakthroughs in fairness, it is crucial to address similar fairness issues in robots. While users are from diverse biological backgrounds, such as race and gender, it is risky to assume that the knowledge and strategies learned from one group of people can apply to others. Discriminatory performance by robots will undermine service quality for particular groups of people, ignore their service requests, and cause offense. For instance, self-driving cars may make racial decisions to harm people while in the "Trolley problem" scenarios to decide who will be harmed in a crash \cite{TheUglyTruth}. It is imperative to mitigate the bias in robots' decision-making processes to improve the inclusivity and quality of robot services.

\begin{figure}[t!]
  \centering
  \includegraphics[width=1\columnwidth]{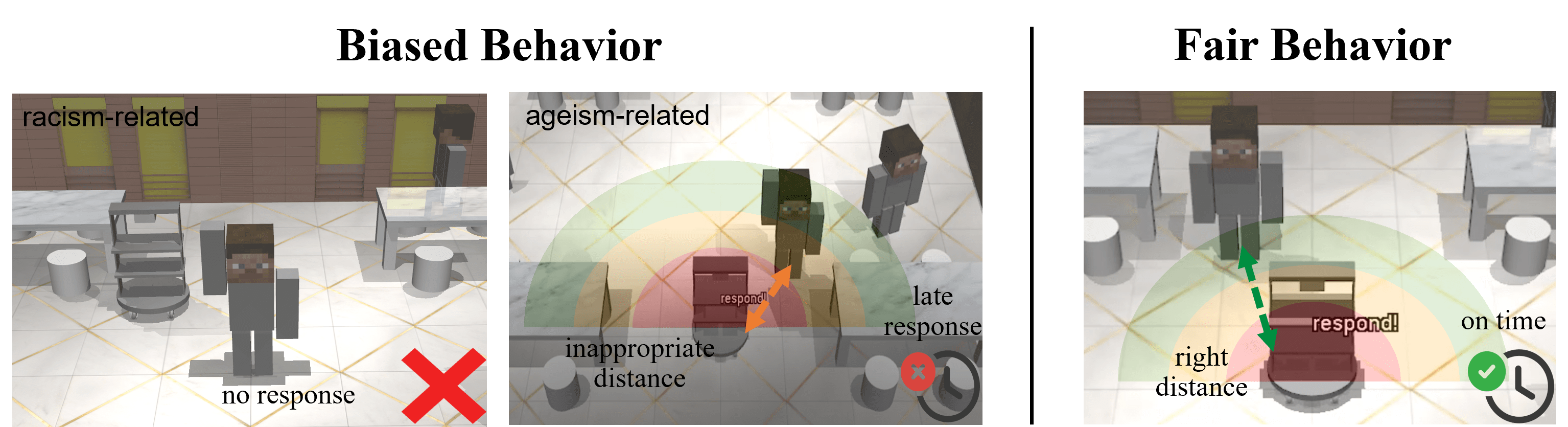}
  \caption{The illustration of robot biased behaviors in the restaurant environment. Our method can detect those biased behaviors and correct them using bias detection guidance.}
  \label{fig:Illustration_robot_bias}
\end{figure}

To mitigate fairness issues in human-robot interactions, our research proposes a self-reflective mechanism -- Fairness-Sensitive Policy Gradient Reinforcement Learning (FSPGRL), to help robots to self-identify biased behaviors during interactions with humans. FSPGRL identifies and mitigates bias in robot learning by examining abnormal updates along specific gradients and updating the policy network to enable fair decision-making. As illustrated in Fig.\ref{fig:Illustration_robot_bias}, biased robot performance without and with mitigations are compared. This paper has three main contributions: 

\begin{enumerate}

\item A novel method "bias detection" was proposed based on a knowledge-informed principal component analysis (PCA); PCA abstracts varying behavior status and then extract the bias-related behavior patterns for timely bias detection during robot service. This bias detection method will enable robots to self-reflect their discriminative behaviors without human reminders.

\item A bias mitigation model was proposed using reinforcement algorithms to adjust robot motions and mitigate its bias dynamically. The robot can perceive biased behaviors and make corrections during the learning procedure and elevate the quality of learning. 

\item A novel bias study was designed to investigate abstract altitude-level bias from various and specific behavior sensory observations (e.g., responding distance and priority). The question design, data processing, and metrics for bias identification and evaluation will provide a protocol for the research community for doing psychology-related robotics and AI research, such as fairness, trust, and safety in human-robot/AI interactions.

\end{enumerate}

\section{RELATED WORK}

\subsection{Fairness in AI and Its Inspiration.}

In general idea, fairness is evaluated by comparing differences in treatment between protected and non-protected groups\cite{FairfromPoliticalPhilosophy}. According to the real-world examples of bias in machine learning algorithms, many works proposed methods of fairness definitions and evaluations like Equal Opportunity \cite{hardt2016equality} and Demographic Parity \cite{dwork2012fairness,kusner2017counterfactual}. Data bias in machine learning is represented as the distribution of specific sensitive attributes that are biased or imbalanced \cite{https://doi.org/10.48550/arxiv.2010.04053}. The solutions of data bias tend to remove discrimination from training data or alter the distribution of sensitive variables \cite{KamiranDataPrePros}. For instance, \cite{karkkainenfairface} introduced \textit{FairFace} dataset by changing the distribution of the unbalanced data to eliminate racial bias in the dataset. Model bias focused on mitigating the prediction bias in machine learning. The IBM Research Trusted AI team came up with an open source toolkit \textit{AI Fairness 360} that can examine, and mitigate discrimination in machine learning models like learning fair representations about protected attributes, etc\cite{bellamy2018ai}. Mengnan proposed Representation Neutralization for Fairness (RNF) by neutralizing fairness-sensitive information in an encoder to reduce the correlation between labels and sensitive properties \cite{https://doi.org/10.48550/arxiv.2106.12674}. \cite{https://doi.org/10.48550/arxiv.1906.12005} proposed a Rényi correlation to measure the fairness of the model and use a novel min-max formulation to balance the accuracy and fairness metrics.\cite{FairnessAdversarialPerturbation} proposed a flexible approach that does not need to retrain existing models to achieve fairness by learning to perturb input data to blind deployed models on fairness-related features. The model and data fairness research focused on comparing model performance in various normal and abnormal statuses to identify issues. 

\subsection{Robot Fairness Issues and The Consequence.} 

Learning from the fairness in AI, some research was done to investigate the fairness issues of robotics and its consequences. For example, \cite{10.3389/frobt.2021.650325} pointed out the importance of fairness in robot navigation, that unfair algorithms of robots will ignore people like black men or interrupt the interaction between people, which might cause people dissatisfaction and even undermine human safety. In tasks involving teamwork, like rescue missions, fairness is essential in building trust between robots and teams to improve team effectiveness \cite{WhyCriteria}. People are concerned about how tasks are distributed and whether the outcome is fair. 

Several methods were proposed to eliminate the potential bias of robots in diverse working environments. For example, \cite{10.3389/frobt.2021.650325} proposed a framework called Learning-Relearning to eliminate bias in social robot navigation by learning social context in navigation first and then relearning with model detection when the navigation model makes biased decisions. \cite{https://doi.org/10.48550/arxiv.2103.09233} proposed continual learning as a bias mitigation strategy for facial recognition tasks in robots to balance learning and robustness against changes in data distributions. \cite{ZhuQinyun2018DRLf} introduced a two-phase distributed algorithm to allocate fair resources for a task by selecting a resource requester first and then forming a robot team. However, these methods mainly focused on robot learning procedures and ignored the interaction between humans and robots. Additionally, the research ignored human feelings from reality and lacked discrimination types or only had empirical definitions of bias, lacking flexibility in real-world usages. Therefore, our work actively explores the relationship between robot behaviors and human feedback through human studies; a sensory data exploration method was developed to detect bias autonomously based on abnormal behavior analysis.
 
\section{Method}

\begin{figure}[t]
  \centering
  \includegraphics[width=1.0\columnwidth]{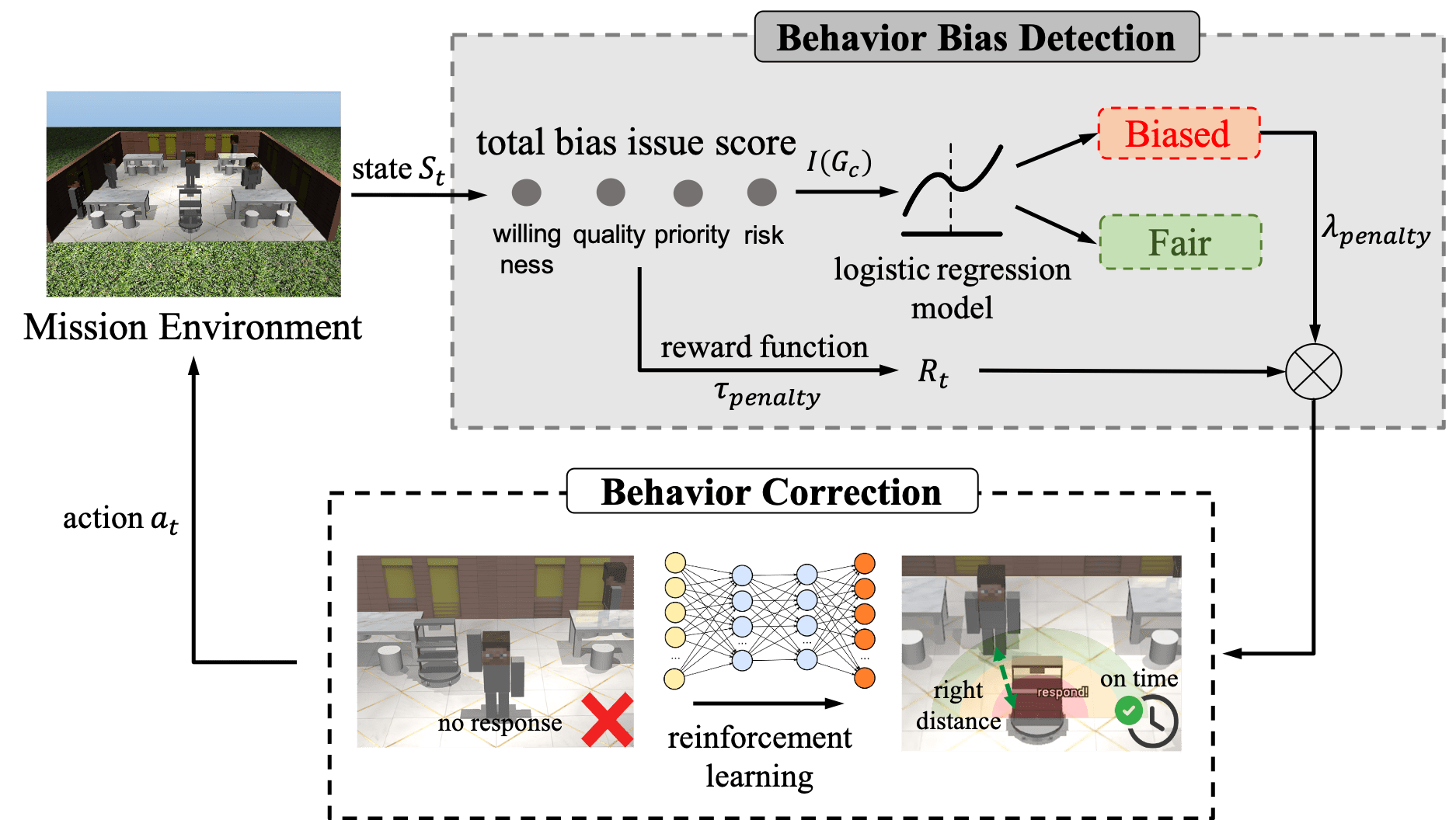}
  \caption{Workflow illustration for bias detection guidance to mitigate robot's discrimination. A pre-trained bias detection model was used to detect if the robot has bias. The robot is trained with a reinforcement learning algorithm and biased behaviors will be corrected during training.}
  \label{fig:Illustration_workflow}
\end{figure}

\textbf{\textit{Preliminaries of FSPGRL.}} The illustration of the workflow of bias detection guidance is shown in Fig. \ref{fig:Illustration_workflow}. Consider a robot serving a requested people $i$ with status $S_t$, where $S_t = (P_i, D_t, t)$. $P_i \in \mathbb{R}^{5}$ denotes sensitive identities of people like race, sex, etc. $t \in \mathbb{R}$ denotes the timestep at status $S_t$. $D_t \in \mathbb{R}$ denotes the distance between the robot and requested people at $t$. $A_t \in \mathbb{R}^{2}$ denotes robot reaction \{\textit{walk, respond}\} at $t$. $R_t \in \mathbb{R}$ denotes the reward of the robot at $t$. 

\subsection{Bias Issues}

We novelly propose the definition of four bias issues: \textit{"Willingness Issue", "Priority Issue", "Quality Issue" and "Risk Issue"}. Let $g_c$ be the group of people that have different identities and $R$ be the behaviors of the robot, the general definition of the issue could be:

\begin{small}
\begin{equation}
d(g_c) = \lvert Pr(R=r \mid P \in g_c)-Pr(R=r \mid P \notin g_c) \rvert
\end{equation}
\end{small}

\noindent\textit{\textbf{Definition 1 (Willingness Issue)}}:
The \textit{Willingness Issue} $d_w \bigl(g_c\bigr)$ describes the different possibilities that robots fail to respond toward the group of people with and without sensitive identities. Based on \cite{pmlr-v81-chouldechova18a}, the ignorance of disadvantaged groups harms their rights and would result in poor user experience. \\
\textit{\textbf{Definition 2 (Quality Issue)}}: 
The \textit{Quality Issue} $d_q \bigl(g_c\bigr)$ describes the different possibilities that robots respond in an inappropriate position toward the group of people with and without sensitive identities. This issue developed based on the navigation robots should avoid an uncomfortable distance towards people \cite{10.3389/frobt.2021.650325}, which will lead to people distrusting robot services and wasting extra time for service compensation.\\
\textit{\textbf{Definition 3 (Priority Issue)}}: 
The \textit{Priority Issue} $d_p \bigl(g_c\bigr)$ describes the different possibilities that robots with overtime response toward the group of people with and without sensitive identities. This issue is inspired by \cite{doi:10.1086/261787} that Pakistan sellers have a delayed discriminatory service. It will lead to people feeling wronged and trigger social criticism. \\
\textit{\textbf{Definition 4 (Risk Issue)}}: 
The \textit{Risk Issue} $d_r \bigl(g_c\bigr)$ describes the different possibilities that robots have a risky distance toward the group of people with and without sensitive identities. In \cite{10.3389/frobt.2021.650325}, It is likely to bring about unsafe collisions between robots and humans.

With $S_t$ generated by the mission environment, bias detection module will calculate the average value of four bias issue score $I(G_c) \in \mathbb{R}^{4}$ to describe the total bias issue score:

\begin{small}
\begin{equation}
I(G_c) = \frac{1}{|G_c|} \sum_{g_{ci} \in G_c} \bigl[d_w(g_{ci}), d_q(g_{ci}), d_p(g_{ci}), d_r(g_{ci})\bigr]
\end{equation}
\end{small}

\subsection{Bias Detection}

In this section, we aim to identify robot discrimination based on its implicit behaviors and attitudes during serving people. We identified bias through a more comprehensive method hybrid with human experience, instead of only using individual thresholds of single abnormal behavior. First, we innovatively use a human study to learn about the human judgment of the biased result denoted as $y$ and collect the dataset denoted as $D$. Second, we use Principal component analysis (PCA)\cite{doi:10.1098/rsta.2015.0202} to extract the implicit proportion of the issue score $I(g_c)$ denoted as $PCA(I(g_c)) \in \mathbb{R}^{3}$ from human study and visualize the behavior of robots via t-SNE. PCA is a method of extracting the main component of features and reducing the dimension of input features. Third, With $y$ and $D$, a logistic regression model was trained to make robot behavior classification. The logistic regression model can learn the experience from human study and judge the discrimination of robot's behavior. The general loss function of logistic regression is defined as:

\begin{small}  
\begin{equation}
\begin{split}
\sum_{(P(I(g_c)), y) \in D}-y \log \left(y^{\prime}\right)-(1-y) \log \left(1-y^{\prime}\right)
\end{split}
\end{equation}
\end{small}

\noindent where $y^{\prime}$ is the predicted labels from model.

The reward penalty parameter $\tau_{penalty} \in \mathbb{R}^{4}$ is weighted and based on different biased issues. we presented following strategies to generate $\tau_{penalty}$:

1). When risk issues appear, human-robot interaction has more possibility to hurt sensitive identities and lead to permanent robots prohibition. Therefore, this kind of bias can be set in the first class $\tau_{penalty}$;

2). When willingness issues appear, the robot is more likely to ignore the sensitive identities which result in mission failures. Therefore such bias can be set in the second class $\tau_{penalty}$;

By utilizing the $\tau_{penalty}$, we can manually adjust the tolerance for various bias issues, allowing for hierarchical reward feeding to the correction section. The reward function is shown as:

\begin{small}
\begin{equation}
R_t = \tau_{penalty} \sum (1 - I(g_c))
\label{con:reward_function1}
\end{equation}
\end{small}

In addition, if the robot was detected as biased, the reward should be lower than normal. To help the robot distinguish between fair and unfair behaviors, we optimized the reward function to guide it. Overall penalty $\lambda_{penalty} \in \mathbb{R}$ is given if the robot was detected as biased by the bias detection model. 

\begin{small}
\begin{equation}
\begin{split}
R_t = \left \{\begin{array}{ll}
\lambda_{penalty} \times  R_t & \text { if detected as biased} \\
R_t & \text { otherwise } \\
\end{array}\right.\end{split} \label{con:reward_function2}
\end{equation}
\end{small}

\subsection{Bias Mitigation}

In this section, the robot is expected to learn what behaviors would be fairer and correct its action during serving. It needs a complicated learning algorithm to achieve the goal. Therefore, we use a vanilla version of the policy gradient method called REINFORCE algorithm \cite{williams1992simple} and Proximal Policy Optimization(PPO)\cite{PPO} to learn the task and eliminate bias in the behavior correction module. Robot has an Actor-Critic network denoted as $A(S_t), C(S_t, a_t)$. PPO takes a balance between ease of implementation and sample complexity. For each training epoch, robot will serve $T=30$ people and collect the set of behaviors $\{a_1, a_2, ..., a_t\}$ by policy $\pi_{\theta_{\mathrm{old}}}\left(a_{t} \mid s_{t}\right)$ then compute the reward $\{r_1, r_2, ..., r_t\}$ base on formulation (\ref{con:reward_function1}) (\ref{con:reward_function2}). An advantage estimate $\hat{A}_{t}$ of people $t$ was used to evaluate current policy:

\begin{small}
\begin{equation}
\begin{split}
-C(S_t, a_t)+r_{t}+\gamma r_{t+1}+\cdots+\gamma^{T-t+1} r_{T-1}+\gamma^{T-t} C(S_T, a_T)
\end{split}
\end{equation}
\end{small}
PPO maximizes the objective via

\begin{small}
\begin{equation}
\begin{split}
L^{CLIP}(\theta)=\hat{\mathbb{E}}_{t}\left[\min \left(r_{t}(\theta) \hat{A}_{t}, \operatorname{clip}\left(r_{t}(\theta), 1-\epsilon, 1+\epsilon\right) \hat{A}_{t}\right)\right]
\end{split}
\end{equation}
\end{small}

where $r_{t}(\theta)$ is the probability ratio:

\begin{small}
\begin{equation}
\begin{split}
r_{t}(\theta)=\frac{\pi_{\theta}\left(a_{t} \mid S_{t}\right)}{\pi_{\theta_{\mathrm{old}}}\left(a_{t} \mid S_{t}\right)}
\end{split}
\end{equation}
\end{small}

$\pi_{\theta}$ is the stochastic policy strategies. $\gamma, \epsilon$ are the hyperparameters. $\operatorname{clip}\left(r_{t}(\theta), 1-\epsilon, 1+\epsilon\right)$ clips the probability ratio in the interval $[1-\epsilon, 1+\epsilon]$.

\section{Evaluation}
To evaluate robots' biased behavior and their discrimination against humans, a simulated physic-based restaurant environment was designed. The following aspects were validated: (i)The effectiveness of bias detection. (ii) The effectiveness of bias detection guidance in reducing robot discrimination during learning and testing. A user study with 24 human volunteers participated in robot discrimination evaluation and 1,000 preference data were simulated. Totally 9,000 interaction demos were recorded and analyzed for comparison.

\begin{figure}[h!]
  \centering
  \includegraphics[width=1.0\columnwidth]{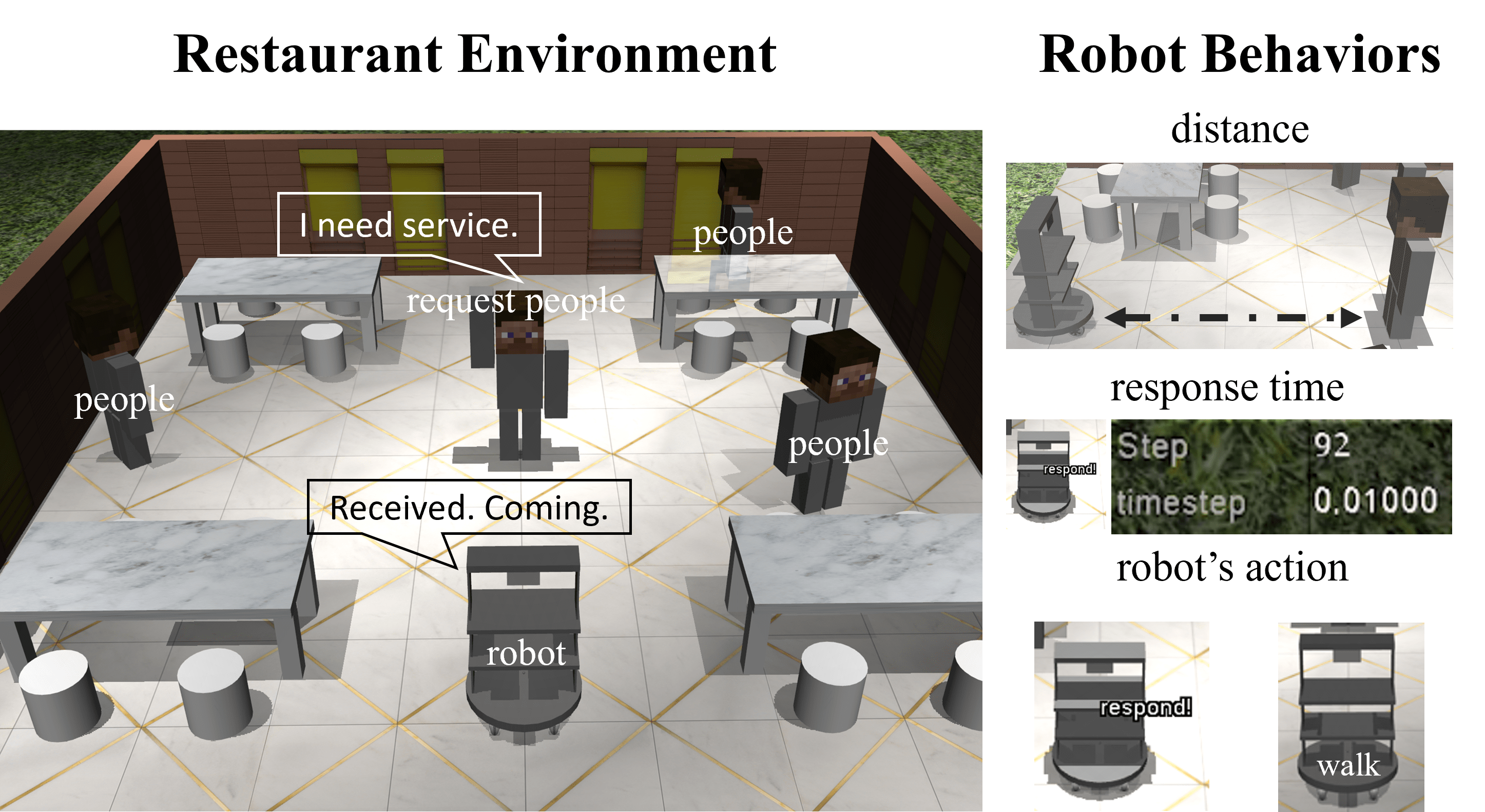}
  \caption{Illustration of the simulated restaurant environment. It contains a robot and a group of people. People with different sensitive identities can raise their hands to indicate they need services from the robot. The robot serves people in the environment.}
  \label{fig:Illustration_env}
\end{figure}

\subsection{Experiment Setting} \label{section: Experiment Setting}
Fig. \ref{fig:Illustration_env} shows the mission environment "A robot is serving people in a restaurant". The robot was designed to deliver food to people and also to assist people if they request a service. The robot has the ability to walk and respond to people. It contains a distance sensor to detect the distance of people and a timer to record response time. There were four types of discrimination: \textit{Racism}, \textit{Sexism}, \textit{Ageism}, and \textit{Ableism}. Inappropriate distance was set as $D_t \notin \mathrm{[1m, 1.5m]}$. Late response time was set as $t > \mathrm{200 \, steps}$, where the step is the refresh step in MUJOCO. The risky distance was set as $D_t < \mathrm{0.5m}$.  $\tau_{penalty}$ was set to [1,1,1,1] and $\lambda_{penalty}$ was set to 0.5. 

Based on the fairness dataset\cite{karkkainenfairface}, we discuss the sensitive identities from five perspectives: race, gender, age, the status of disability, and skin color. The race includes \textit{White}, \textit{Black}, \textit{American Indian}, \textit{Asian}, \textit{Native Hawaiian}, and \textit{Other race}. The gender includes \textit{Female}, \textit{Male}, and \textit{Other}. The age was divided into \textit{Child}, \textit{Teenager}, \textit{Adult}, \textit{Middle Aged}, and \textit{Elder}. The disability of people was set as \textit{Yes} or \textit{No}. The skin color was set as \textit{Type \uppercase\expandafter{\romannumeral1}} to \textit{Type \uppercase\expandafter{\romannumeral6}} based on \textit{Fitzpatrick scale} \cite{Sachdeva2009FitzpatrickST}. 

\begin{figure}[t!]
  \centering
  \includegraphics[width=0.8\columnwidth]{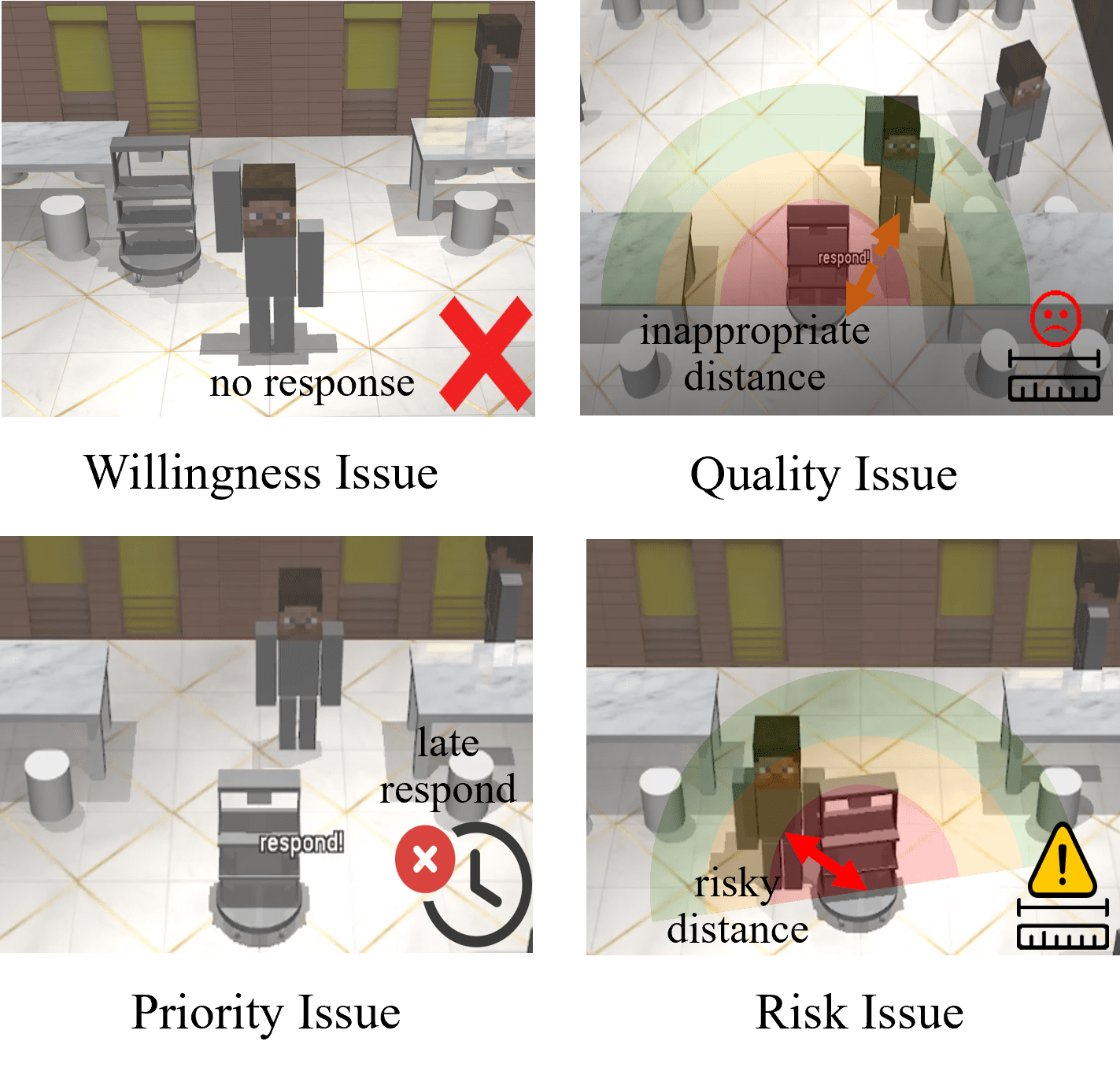}
  \caption{The illustration of robot biased behaviors. Willingness issue is represented as the robot does not respond to requested people. Quality issue is represented as the robot responds but at a close distance. Priority issue is represented as the robot responds late. Risk issue is represented as the robot at a risky distance towards people in any step.}
  \label{fig:Robot_bias_bahavior}
\end{figure}

There is a single task in the scenario: human interactive task. The robot needs to provide continuous service to a group of people. One person will request service in each episode, and the robot needs to respond. If the robot completes the task or is beyond the time limit, the episode ends and goes to the next episode. People's identities and positions are randomly generalized in each episode.

\subsection{Result Analysis}

\begin{figure}[h!]
  \centering
  \includegraphics[width=1\columnwidth]{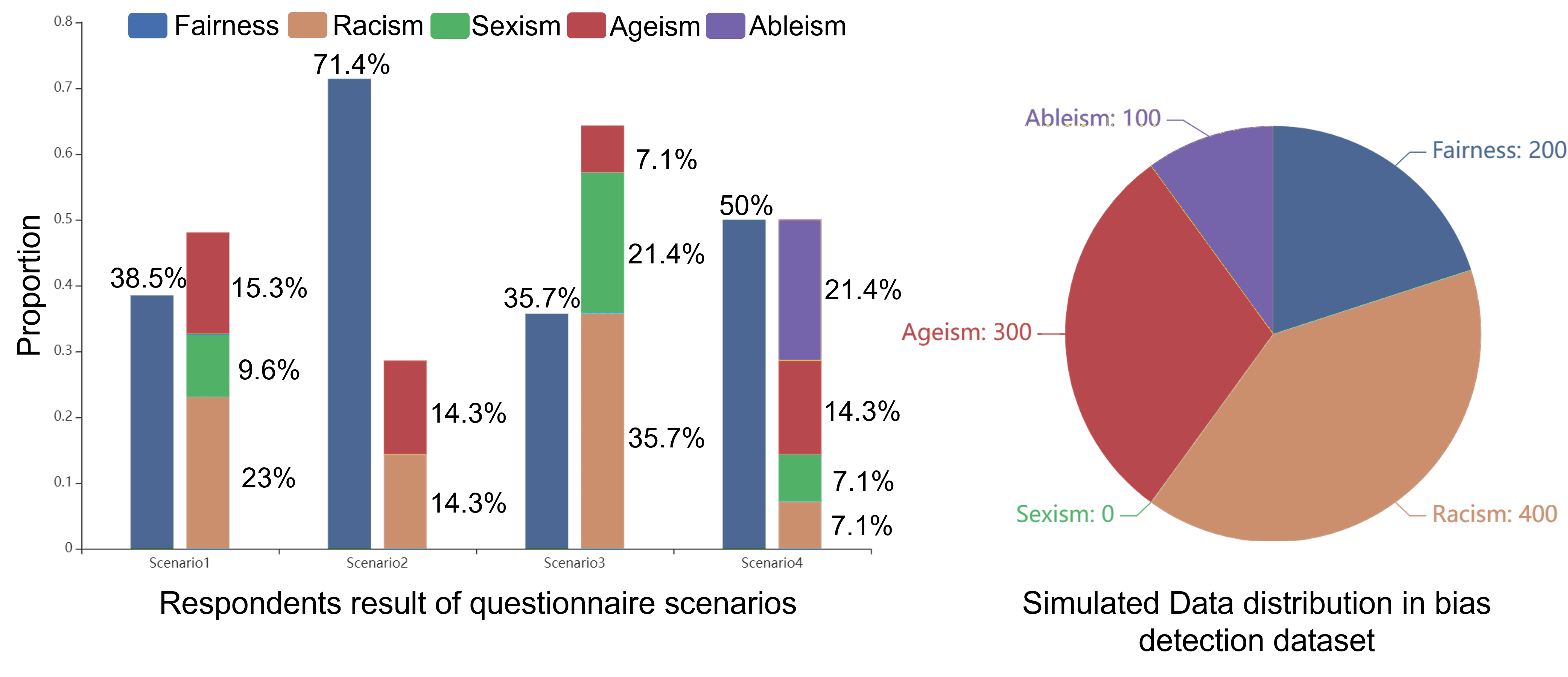}
  \caption{Result from the questionnaire. The left chart is the distribution of respondents' results of the scenarios. The right chart is the dataset distribution based on the simulated data.}
  \label{fig:Questionnaire_result}
\end{figure}

\indent \textbf{User study.} A questionnaire was set up by collecting the robot interaction data in our environment. We ensure the diversity of volunteers in order to get a relatively fair analysis. In the questionnaire, volunteers were asked some questions like which issue is most serious and what they think about the robot's behaviors. The result of respondents and the simulated data distribution is shown in Fig. \ref{fig:Questionnaire_result}. Racism is strongly related to Willingness Issue with 35.7\% of the volunteers thinking the robot responded badly to the Black race. 21.4\% of the volunteers chose ableism in scenario 4 since it is closed to people when the customer is disabled. 

\begin{figure}[h!]
  \centering
  \includegraphics[width=0.8\columnwidth]{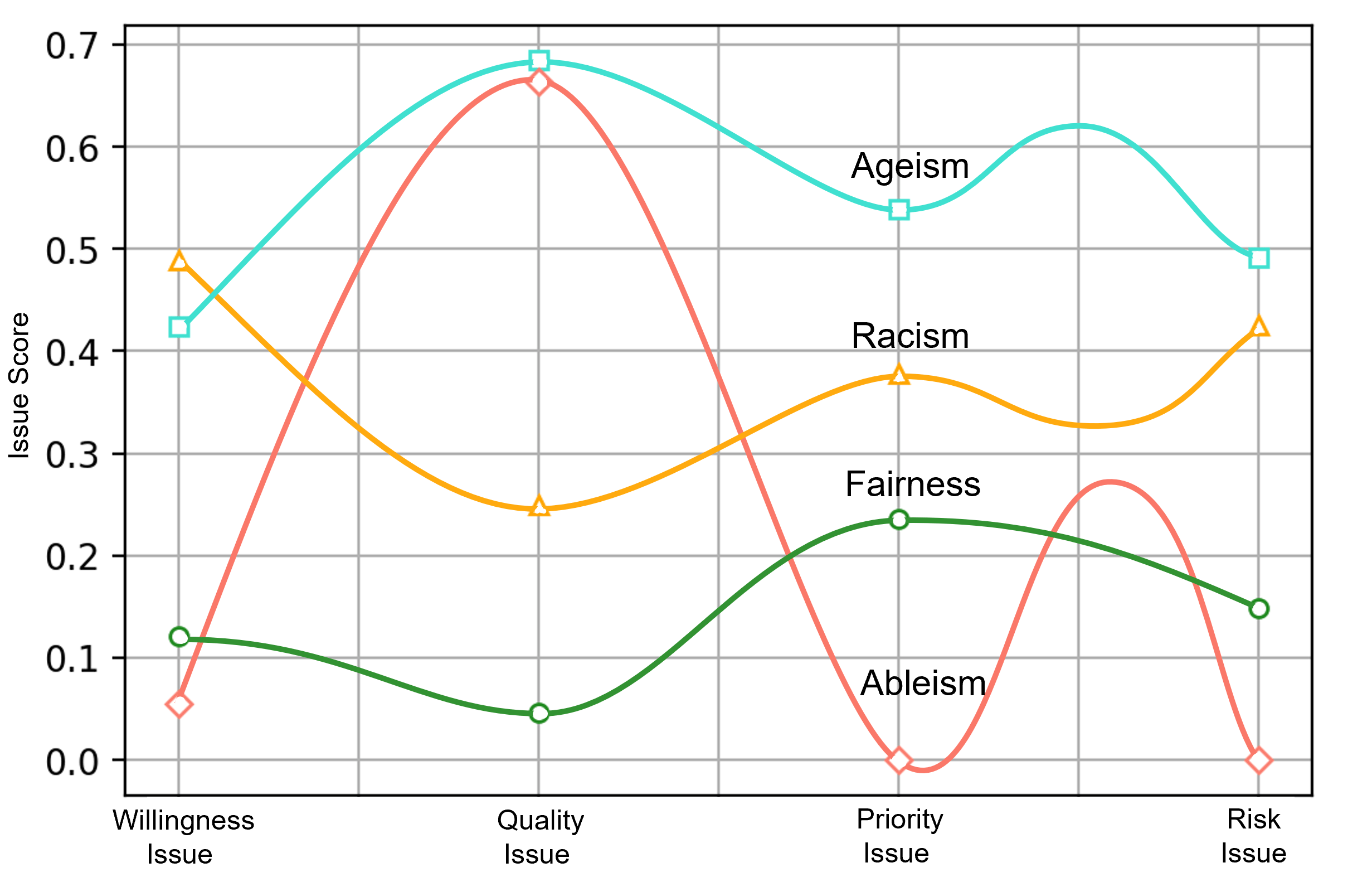}
  \caption{Characteristic of discrimination learning from user study.}
  \label{fig:Bias_charact}
\end{figure}

\begin{figure}[h!]
  \centering
  \includegraphics[width=1\columnwidth]{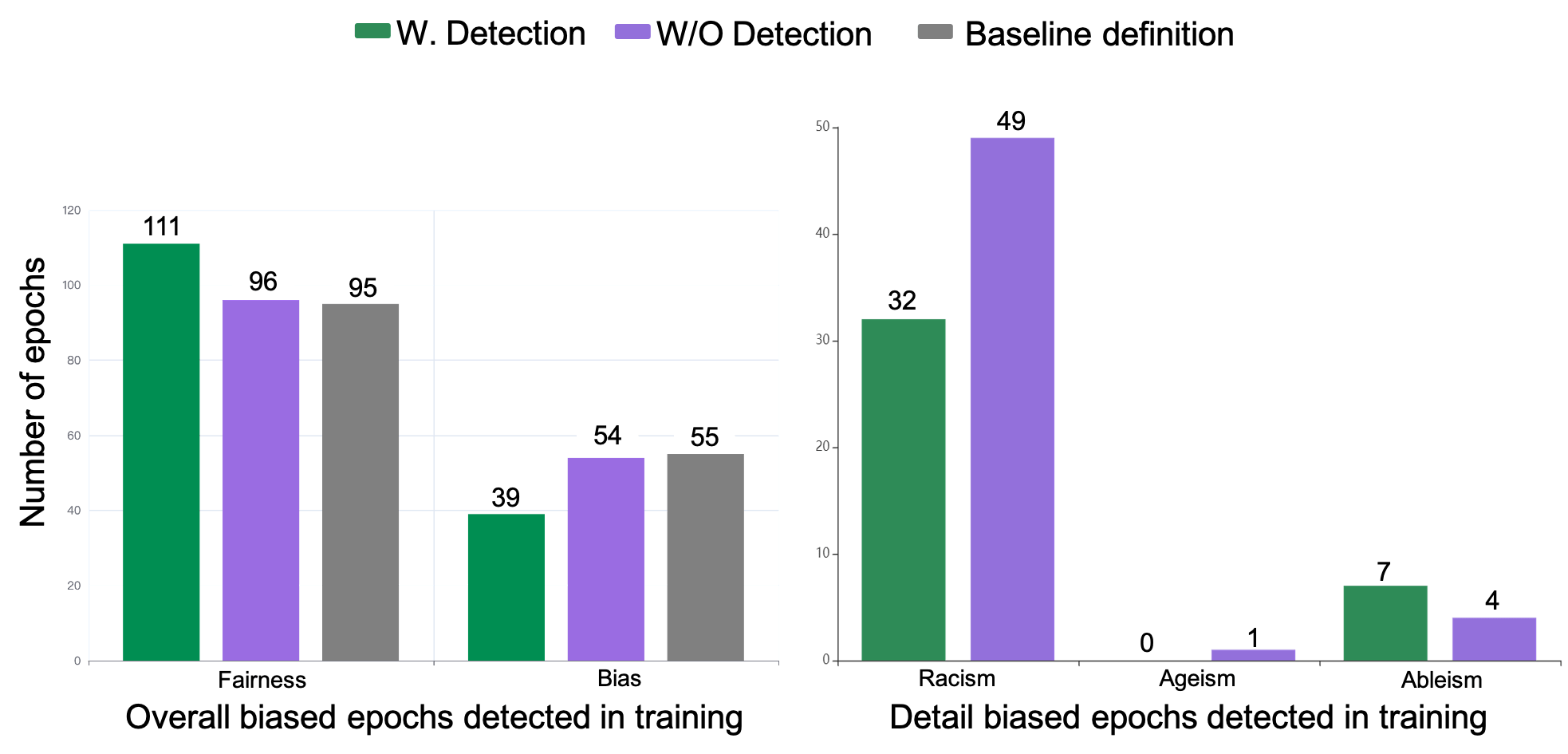}
  \caption{Robot learning performance comparison. The robot with bias detection guidance (green) detected 39 biased epochs and the robot without guidance (purple) detected 54 biased epochs by bias detection model during the learning procedure.}
  \label{fig:Bias_detection_comparsion}
\end{figure}

\indent \textbf{Bias detection validation.} In this part, we built a bias detection model by the simulated data from the questionnaire. Since sexism was not included in the simulated data, our further analysis will not contain it. Visualized bias characteristic is shown in Fig. \ref{fig:Bias_charact}. We observed that each discrimination has its own signatures considering 4 bias issues learned from user study. Results indicate racism is strongly related to the Willingness Issue with an issue score of 0.5, represented by vector [0.49, 0.25, 0.38, 0.43] corresponding with four bias issues; ageism to Quality, Priority, and Risk Issues, represented by vector [0.42, 0.68, 0.54, 0.49]; and ableism to Quality with a score of nearly 0.7, represented by vector [0.05, 0.66, 0, 0]. In the subsequent experiment in Fig. \ref{fig:Bias_detection_comparsion}, we demonstrated the validity of our bias detection model with 98\% detection accuracy compared with manual detection and model detection on the overall bias. The grey bar indicates the number of biased epochs exceeding the manually set bias threshold (0.5).

\begin{figure}[h!]
  \centering
  \includegraphics[width=0.80\columnwidth]{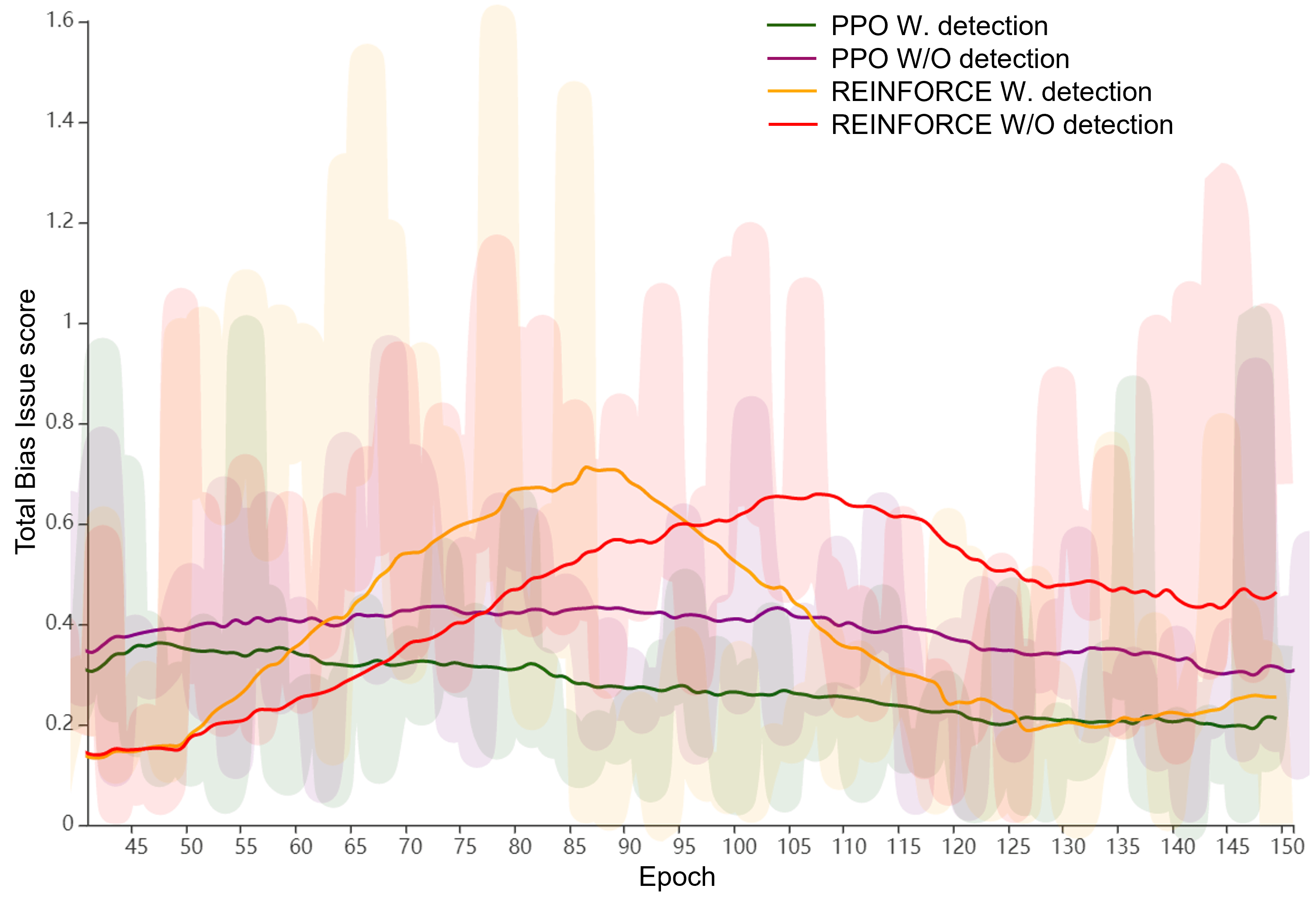}
  \caption{Total bias issue score comparison during learning. The background color area indicates the total bias issue score of each epoch. Color lines indicate the 40 average epochs of the total bias issue score.}
  \label{fig:Total_issue_score}
\end{figure}

\begin{figure}[h!]
  \centering
  \includegraphics[width=1\columnwidth]{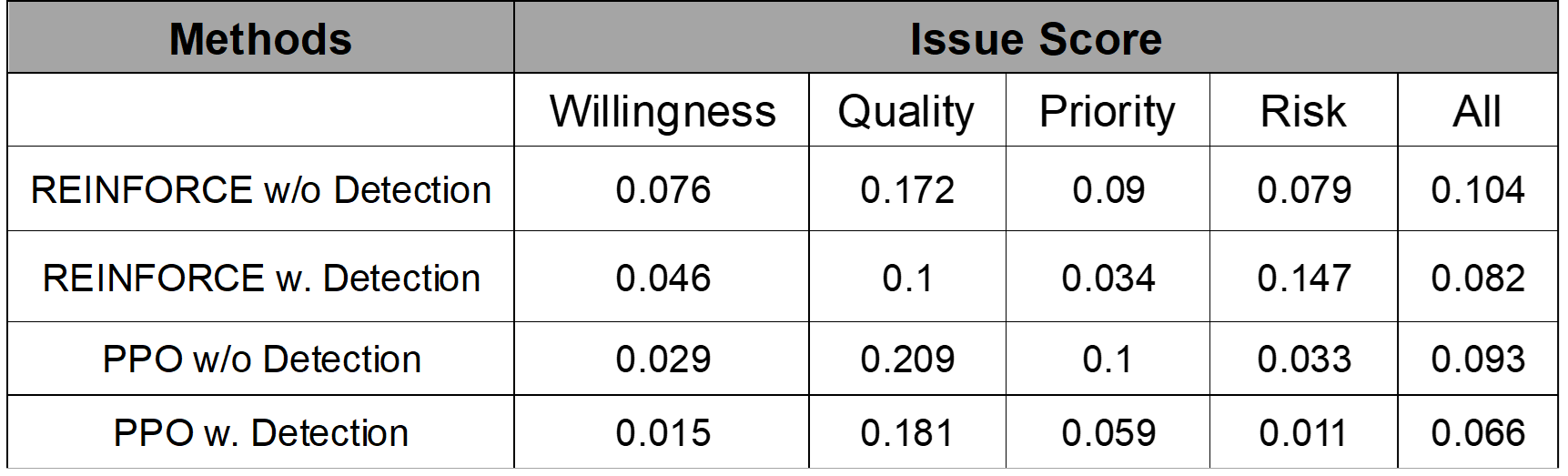}
  \caption{Result of bias issue score comparison in the test dataset.}
  \label{fig:table_performance}
\end{figure}

\begin{figure}[h!]
  \centering
  \includegraphics[width=0.80\columnwidth]{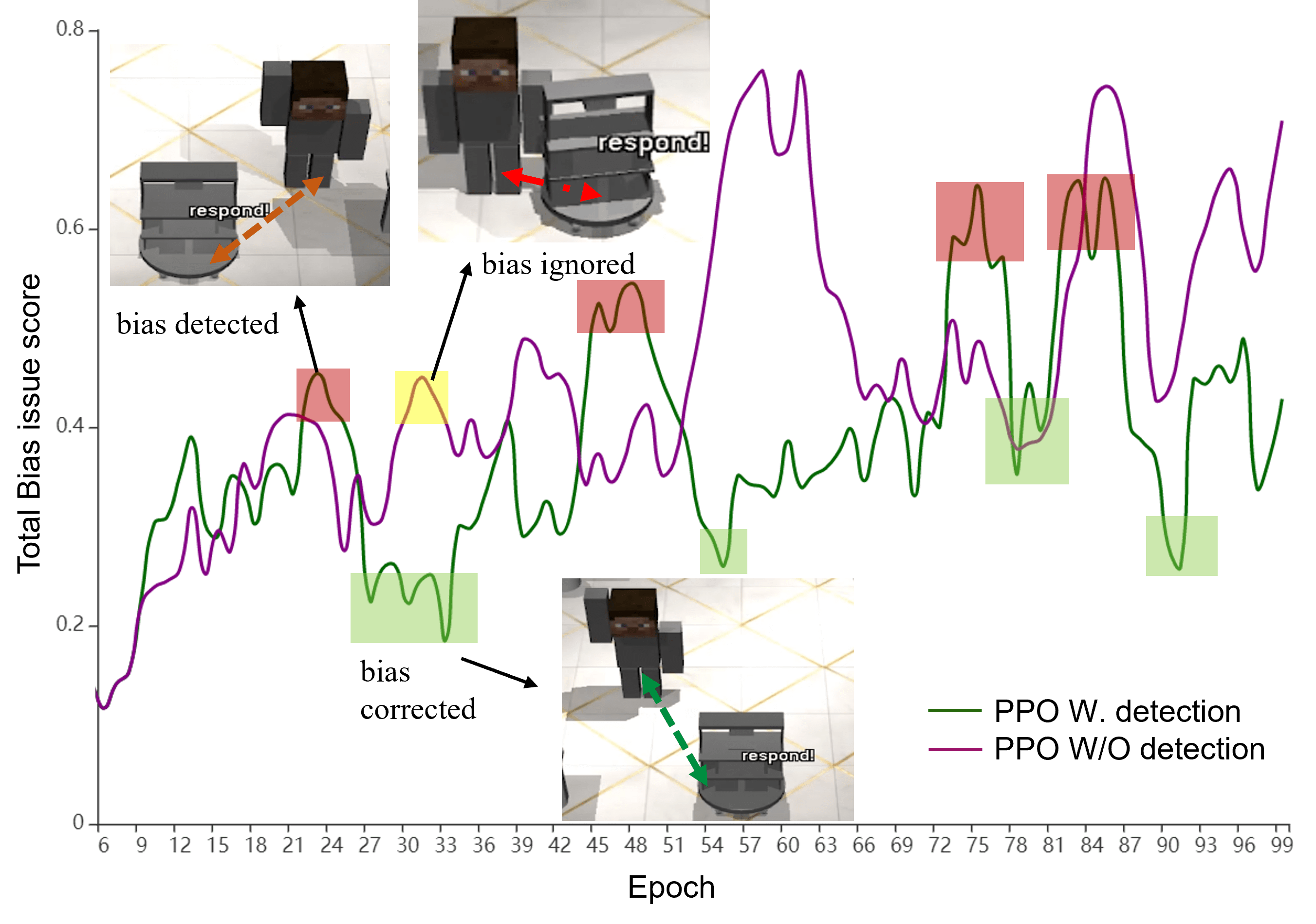}
  \caption{The illustration of bias detection guidance. The red areas indicate the epochs that are detected as biased. The green areas show that with bias detection guidance, the robot perceives biased behaviors and corrects it. The yellow area shows that without bias detection guidance, the robot can not correct its biased behavior and has a worse performance further.}
  \label{fig:Bias_detection_guidance_detail}
\end{figure}

\indent \textbf{Bias detection guidance validation.} The result of the experiment was shown in Fig. \ref{fig:Total_issue_score}. The method with bias detection guidance can reach lower total bias issue scores to mitigate bias. The detailed issue score of the test dataset is shown in Fig. \ref{fig:table_performance}. With bias detection guidance, both REINFORCE and PPO can reach a significantly more than 22\% lower total bias issue score than without it. As shown in Fig. \ref{fig:Bias_detection_comparsion}, the robot successfully achieved the suppression of bias and enhanced the quality of learning during the learning process, which validates the effectiveness of bias detection guidance. PPO with bias detection guidance reached 27.8\% more fairness epochs with bias detection guidance than the PPO without it. Once the bias detection model identifies the bias in training, it will try to eliminate bias via penalty. The detail of bias detection guidance is shown in Fig. \ref{fig:Bias_detection_guidance_detail}. With our method, robots can identify biased behaviors and correct them rapidly, so the total bias issue score can be confined to a small interval. Without bias detection guidance, the total bias issue score exhibited higher volatility and increment.

\begin{figure}[h!]
  \centering
  \includegraphics[width=1.0\columnwidth]{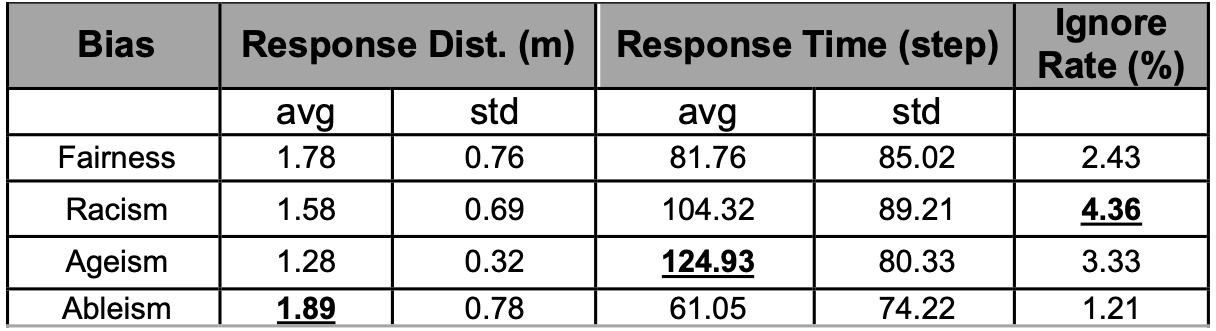}
  \caption{Result of the response distance, response time and ignore rate of four types of behavior. Data in bold and underline indicate the maximum value of columns.}
  \label{fig:table_character}
\end{figure}

\begin{figure}[h!]
  \centering
  \includegraphics[width=0.65\columnwidth]{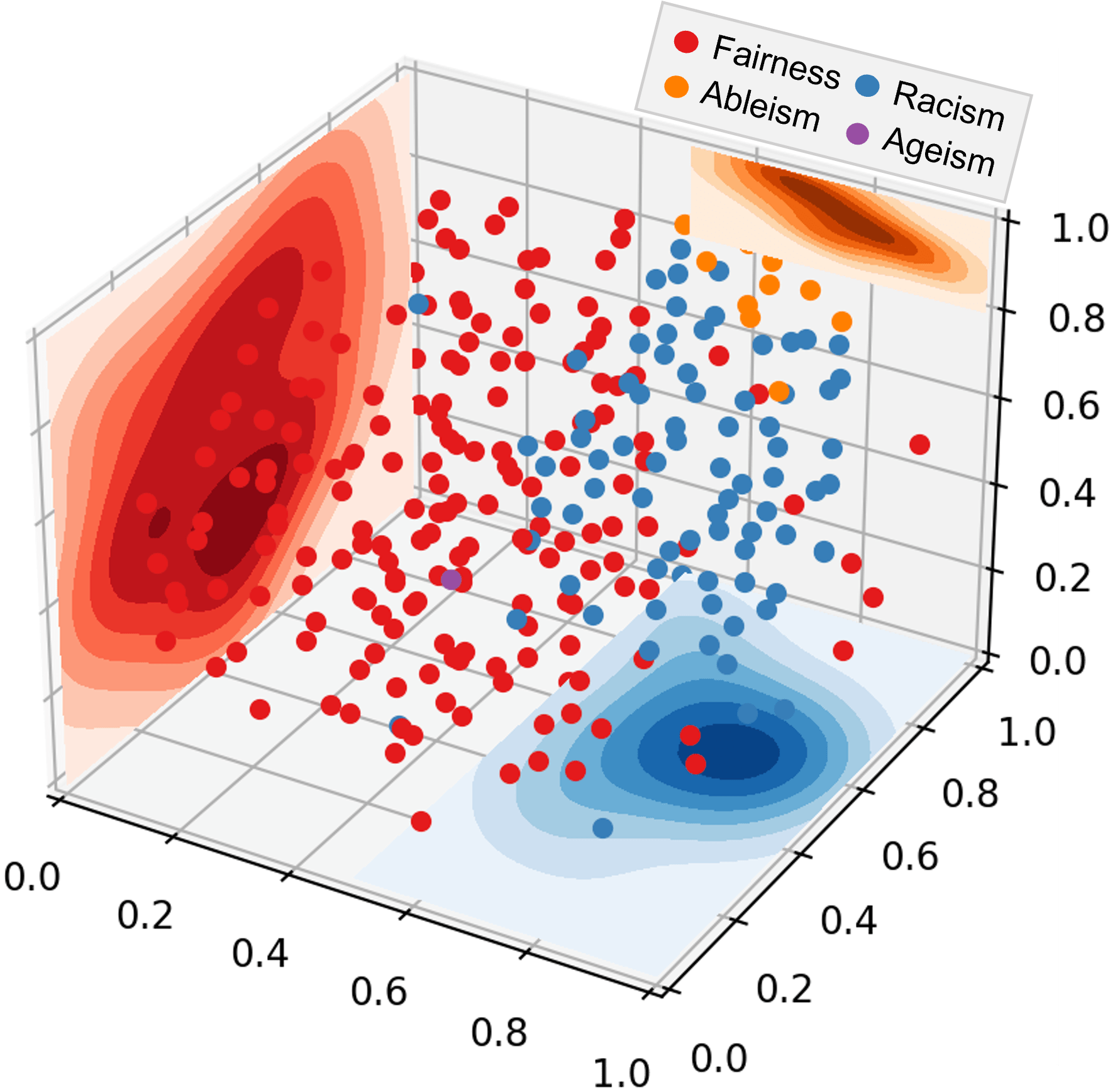}
  \caption{t-SNE of dataset distribution. The density distribution of discrimination on 3 planes indicates a strong correlation between different metrics and bias.}
  \label{fig:tsne}
\end{figure}

\indent \textbf{Biased behavior evaluation.} Our results revealed that robot has different action patterns of robots with respect to biased behaviors. Each type of bias has different emphases on bias issues. Fig. \ref{fig:table_character} displayed the data disparities of four types of bias. Ableism behavior has the farthest average response distance at 1.89 than other behavior, while ageism behavior has the longest average response time at 124.93. Racism behavior has the largest ignore rate at 4.36 compared to other behavior. Fig. \ref{fig:tsne} illustrates the distinctive distribution of four types of discrimination. In conclusion, racism is represented by robots refusing to provide services to them. Ageism is expressed as providing poor service resulting in long wait times. Ableism is exposed as an inappropriate social distance when robots serve the disabled. These findings highlight the similarities between robots' biased representation and humans' biased experience in Fig. \ref{fig:Bias_charact}. 

\section{Conclusion and Future Work}
In this work, We measured bias from more comprehensive perspectives by incorporating human experience and introduced a bias detection guidance method to effectively mitigate robot discrimination against people and achieved robot self-awareness of bias. To validate our method's effectiveness, a human interactive task was deployed in a restaurant environment; a user study was done to identify the robot's biased behaviors. The superiority of our method for robot bias detection and correction was validated by the efficiency of detection accuracy, total bias issue score reduction, and the suppression of bias during training. In the future, bias evaluation and mitigation methods can be further improved with more sensitive detection methods and more dimensions to measure robot behavior; complicated situations like multi-robot cooperation that are closer to real life can be investigated.

\addtolength{\textheight}{0cm}   


\bibliographystyle{unsrt2authabbrvpp}
\bibliography{root}

\end{document}